\pdfoutput=1

\documentclass[11pt]{article}

\usepackage[]{emnlp2021}

\usepackage{times}
\usepackage{latexsym}
\usepackage{xspace}
\usepackage{multicol}
\usepackage{multirow}
\usepackage{graphicx}
\usepackage{nicefrac}
\usepackage[T1]{fontenc}

\usepackage[utf8]{inputenc}

\usepackage{microtype}

\usepackage{kotex}
\usepackage[para]{threeparttable}
\usepackage{booktabs}

\newcommand{\modelname}{HyperCLOVA\xspace}

\title{What Changes Can Large-scale Language Models Bring?\\ Intensive Study on HyperCLOVA: Billions-scale Korean Generative Pretrained Transformers}

\author{Boseop Kim\thanks{\ \ Equal contribution.}\ $^{,1}$\ \ \  HyoungSeok Kim$^{*,1}$\ \ \  Sang-Woo Lee$^{*,1,2}$\ \ \  Gichang Lee$^{1}$\\
\textbf{Donghyun Kwak$^{1}$\ \ \  Dong Hyeon Jeon$^{3}$\ \ \  Sunghyun Park$^{4}$\ \ \  Sungju Kim$^{1,3}$} \\
\textbf{Seonhoon Kim$^{3}$\ \ \  Dongpil Seo$^{1}$\ \ \  Heungsub Lee$^{1}$\ \ \  Minyoung Jeong$^{1}$\ \ \  Sungjae Lee$^{1}$}\\ 
\textbf{Minsub Kim$^{1}$\ \ \  Suk Hyun Ko$^{1}$\ \ \  Seokhun Kim$^{1}$\ \ \  Taeyong Park$^{1}$\ \ \ Jinuk Kim$^{1}$} \\
\textbf{Soyoung Kang$^{1}$\ \ \  Na-Hyeon Ryu$^{1}$\ \ \  Kang Min Yoo$^{1,2}$\ \ \  Minsuk Chang$^{2}$\ \ \  Soobin Suh$^{1,3}$}\\
\textbf{Sookyo In$^{1,3}$\ \ \  Jinseong Park$^{1,3}$\ \ \  Kyungduk Kim$^{1,3}$\ \ \  Hiun Kim$^{1}$\ \ \  Jisu Jeong$^{1,2}$}\\
\textbf{Yong Goo Yeo$^{1}$\ \ \  Donghoon Ham$^{1}$\ \ \  Dongju Park$^{1}$\ \ \  Min Young Lee$^{1}$\ \ \  Jaewook Kang$^{1}$}\\
\textbf{Inho Kang$^{1,3}$\ \ \  Jung-Woo Ha$^{1,2}$\ \ \  Woomyoung Park$^{1}$\ \ \  Nako Sung$^{1}$}\\

\\
NAVER CLOVA$^{1}$\ \ \  NAVER AI Lab$^{2}$\ \ \  NAVER Search$^{3}$\ \ \ Search Solutions, Inc.$^{4}$}

\begin{document}
\maketitle
\begin{abstract}
GPT-3 shows remarkable in-context learning ability of large-scale language models (LMs) trained on hundreds of billion scale data.
Here we address some remaining issues less reported by the GPT-3 paper, such as a non-English LM, the performances of different sized models, and the effect of recently introduced prompt optimization on in-context learning.
To achieve this, we introduce \modelname{}, a Korean variant of 82B GPT-3 trained on a Korean-centric corpus of 560B tokens.
Enhanced by our Korean-specific tokenization, \modelname{} with our training configuration shows state-of-the-art in-context zero-shot and few-shot learning performances on various downstream tasks in Korean. Also, we show the performance benefits of prompt-based learning and demonstrate how it can be integrated into the prompt engineering pipeline. Then we discuss the possibility of materializing the No Code AI paradigm by providing AI prototyping capabilities to non-experts of ML by introducing \modelname{} studio, an interactive prompt engineering interface. Lastly, we demonstrate the potential of our methods with three successful in-house applications.
\end{abstract}

\section{Introduction}
Due to its remarkable zero-shot and few-shot performances, GPT-3's in-context learning has gained significant attention in the AI community \cite{brown2020language}. In the in-context learning approach, discrete prompts that consist of a natural language task description and few-shot examples control large-scale language models (LMs) to infer predictions for the target task. Using OpenAI's GPT-3, studies have proposed methods that can further boost in-context learning performance of GPT-3 \cite{zhao2021calibrate,liu2021makes}.
Recently, prompt-based learning methods have been reported to improve the performances of BERT, GPT-3, and T5 without any parameter updates of the main model~\cite{liu2021gpt,lester2021power,shin2020autoprompt}.

We consider the three following practical issues of using GPT-3. First, the language composition of the training corpus is heavily skewed towards English with 92.7\%. This makes it difficult to apply it to tasks in other languages. We also know little about how to train similar models in another language with different linguistic properties, and where the originally proposed methods will be naturally applied and where they might fail. Second, while it is pragmatic and useful to know the capabilities of various sized models considering the operation costs of using large-scale LMs, we only have access to a thorough analysis of models of 13B and 175B~\cite{brown2020language} but none in between. Lastly, advanced prompt-based learning methods that require backward gradients of inputs, including continuous prompt-based tuning, have not yet been experimented for an in-context large-scale LM learner.

Here we address these issues by introducing a non-English GPT-3 with various parameter sizes and intensively investigating their capabilities on diverse real-world classification and generation tasks under in-context few-shot learning and prompt-based optimization. We introduce a Korean in-context large-scale LM with 82B parameters, i.e., \modelname{}. This is the first discovery on near 100B-scale non-English LM. 
We present the corpus composition of Korean datasets used for \modelname{}, 
and describe how we crawl and refine such data to collect 561B tokens of Korean corpus (\S \ref{subsec:data-decription}). 
We also design a new Korean tokenization method based on the agglutinative property for \modelname{}. We use byte-level BPE \cite{kudo2018sentencepiece} with a morpheme analyzer (\S \ref{subsec:korean-tokenization}). Our results show that such tokenization strategy is important for the performance of downstream tasks in large-scale in-context learning (\S \ref{subsec:effect-of-tokenization}).

We report the state-of-the-art in-context learning performance of our model on Korean datasets in zero and few-shot settings (\S \ref{subsec:in-context-few-shot-learning}). 
In addition, we are the first to discovery the applicability of the continuous prompt-based optimization techniques, such as p-tuning \cite{liu2021gpt}, to large-scale LMs. \modelname{} leveraged by p-tuning achieves outstanding results for both classification and generation tasks. Also, we investigate the effects of p-tuning on two mid-size \modelname{} (\S \ref{subsec:prompt-based-tuning}). 

Subsequently, we illustrate the versatility of operating a single large-scale LM in the AI industry.
Developing an AI product involves heavy collaboration among non-technical experts. This incurs substantial communication overhead because the level of technical abstraction varies across job functions. 

We introduce \modelname{} Studio, an interactive prompt engineering interface which provides GUI and API interfaces like the OpenAI playground\footnote{https://beta.openai.com/}. The interactive interface helps non-experts of ML to easily use \modelname{} for prototyping AI products.
We also share three in-house application scenarios using \modelname{} Studio as novel task environments.
With minimal efforts of a domain expert in these scenarios, \modelname{} presents performances qualitatively comparable to human experts, despite their difficulty in designing their objective function and training data (\S \ref{subsec:discussion-on-in-house-usage}). 

We then discuss how the functionality of \modelname{} Studio can be extended. For example, \modelname{} Studio can provide input gradient functionality, to fine-tune small prompt encoder with few number of instances, thus enabling any user to achieve state-of-the-art performance using \modelname{} (\S \ref{subsec:opportunity}). 
Finally, we discuss the possibility of No/Low Code AI paradigm using \modelname{} Studio, in which one large LM empowers people to create AI systems with no need for training individual deep learning models or collecting and labeling suitable datasets (\S \ref{subsec:no-code-ai-paradigm}). 

Our contributions are summarized as:
\begin{enumerate}
    \item We introduce \modelname{}, a large-scale Korean in-context learning-based LM with nearly 100B parameters, by constructing a large Korean-centric corpus of 560B tokens. 
    \item We discover the effect of language-specific tokenization on large-scale in-context LMs for training corpus of non-English languages.
    \item We explore the zero-shot and few-shot capabilities of mid-size \modelname{} with 39B and 82B parameters and find that prompt-based tuning can enhance the performances, outperforming state-of-the-art models on downstream tasks when backward gradients of inputs are available.
    \item We argue the possibility of realizing No Code AI by designing and applying \modelname{} Studio to our in-house applications. We will release \modelname{} Studio with input gradients, output filters, and knowledge injection.
\end{enumerate}

\section{Previous Work}

\subsection{Prompt Optimization}
Prompt-based approaches involve constructing optimal prompts for language models to best elicit knowledge and maximize prediction performances \cite{radford2019language,brown2020language,schick2020s}.
As the scale of language models grows, the potential of replacing the full fine-tuning paradigm with the prompt-based approach has been reported \cite{reynolds2021prompt, li2021prefix_}, as learning via prompts is efficient regarding time and space complexity. However, language models are highly sensitive to the prompt design, motivating methodologies for optimizing prompts. 

Prompt optimization can be categorized into \emph{discrete} and \emph{continuous} approaches. The discrete approach optimizes directly on the token space \cite{ben2021pada, shin2020autoprompt} and has the advantage of transferability. However, \citet{shin2020autoprompt} showed that the discrete space has poor interpretability and can be suboptimal. These limitations spurred a new direction that aims to optimize prompts in the continuous space. Recent work \cite{li2021prefix_, hambardzumyan2021warp, liu2021gpt, lester2021power} proposed optimizing the contextualized token spaces without fine-tuning the main LM parameters. Notably, \citet{liu2021gpt} found that p-tuning for autoregressive LMs outperforms MLM-based fine-tuning in certain downstream tasks. \citet{lester2021power} further showed that well-optimized prompt-based learning achieves state-of-the-art performance on key benchmarks.

\subsection{Language Models}
Although multilingual language models have been publicly available \citep{devlin2019bert}, language-specific language models are still in demand, as they provide an edge over language-agnostic models \cite{martin2020camembert,nguyen2020phobert,delobelle2020robbert}. However, due to high cost, language-specific language models other than English are limited in availability.

As such, the community has an untapped understanding of non-English in-context learners.
To the best of our knowledge, multilingual in-context learners are not even explored yet, and the research on in-context learners is focused on few major languages.
Recently,  a GPT-like language model trained on Chinese corpora is being actively researched concurrently with our work \cite{zeng2021pangu}. They successfully trained LMs of 2.6B and 13B parameters using a Chinese corpus. They also share their on-going work for training the 207B model, the corresponding infrastructure, and the training techniques.

\section{Pre-training}

\subsection{Data Description}
\label{subsec:data-decription}
The ratio of Korean data for OpenAI GPT-3 is very small, with less than 0.02\% by character count.\footnote{https://github.com/openai/gpt-3/blob/master/dataset\_statistics/languages\_by\_word\_count.csv} 
Therefore, it is crucial to construct a large Korean-centric corpus in advance to training \modelname{}.

The major corpus used for pre-training \modelname{} is listed in Table \ref{table:table1}. 
To build a large-scale corpus comparable to that for training OpenAI GPT-3, we gathered all available text data including user-generated content (UGC) and contents provided by external partners, with no violation of legal issues, from both diverse services of NAVER\footnote{https://www.naver.com/} and external sources.

We refined the datasets and collected a total of 561B tokens as the final corpus. 
The corpus was randomly sampled for pre-training.
Appendix \ref{subsec:data-description} describes the detailed data description and discussion.  
Appendix \ref{subsec:data-cleaning}, \ref{subsec:data-anonymization}, and \ref{subsec:data-postprocessing} thoroughly describe how to clean, anonymize, and preprocess the crawled raw data, respectively. 

\begin{table}[t!]
    \centering
    \small
    \setlength\tabcolsep{4.5pt}
    \begin{threeparttable}
    \begin{tabular}{llr}
        \toprule
        Name& Description & Tokens\\
        \midrule
        Blog & Blog corpus & 273.6B \\
        Cafe & Online community corpus & 83.3B \\
        News & News corpus & 73.8B \\
        Comments & Crawled comments & 41.1B \\
        KiN & Korean QnA website & 27.3B \\
        Modu & Collection of five datasets  & 6.0B \\
        WikiEn, WikiJp & Foreign wikipedia & 5.2B \\
        Others & Other corpus & 51.5B\\
        \midrule
        Total & & 561.8B \\
        \bottomrule
    \end{tabular}
    \end{threeparttable}
    \caption{Descriptions of corpus for \modelname{}.}
    \label{table:table1}
\end{table}

\begin{table}[t!]
    \centering
    \small
    \setlength\tabcolsep{4pt}
    \begin{threeparttable}
    \begin{tabular}{lccccc}
        \toprule
        \# Param & $n_{layers}$ & $d_{model}$ & $n_{heads}$ & $d_{head}$ & $lr$\\
        \midrule
        137M & 12 & 768 & 16 & 48 & 6.0e-4 \\
        350M & 24 & 1024 & 16 & 64 & 3.0e-4\\
        760M & 24 & 1536 & 16 & 96 & 2.5e-4\\
        1.3B & 24 & 2048 & 16 & 128 & 2.0e-4\\
        6.9B & 32 & 4096 & 32 & 128 & 1.2e-4\\
        13B & 40 & 5120 & 40 & 128 & 1.0e-4\\
        39B & 48 & 8192 & 64 & 128 & 0.8e-4\\
        82B & 64 & 10240 & 80 & 128 & 0.6e-4\\
        \bottomrule
    \end{tabular}
    \end{threeparttable}
    \caption{Detailed configuration per size of \modelname{}.}
    \label{table:table11}
\end{table}

\subsection{Model and Learning}
We employ the same transformer decoder architecture as GPT-3 of OpenAI \cite{brown2020language}. Table \ref{table:table11} describes the detailed configurations of different model sizes.
We make our model design similar to GPT-3, and we set near exponential interpolation from 13B to 175B OpenAI GPT-3. In particular, we aim to explore the capability and representation power of the models with mid-size parameters, which have not yet been addressed by other studies on large-scale LMs \cite{brown2020language}, but practically useful in many applications.
These mid-size models can contribute to not only understanding the model properties with several tens of billion parameters, but also practical usages in real-world applications due to their more plausible sizes. 

Our model is based on megatron-LM \cite{shoeybi2019megatron} and trained on the NVIDIA Superpod, which includes 128 strongly clustered DGX servers with 1,024 A100 GPUs. We use AdamW~\cite{loshchilov2018decoupled} with cosine learning rate scheduling and weight decay as an optimizer.  All models use the mini-batch size of 1,024 and the minimum learning rate is 1/10 of the original learning rate. It takes 13.4 days to train a model with 82B parameters with 150B tokens. 
For experiments in Section \ref{sec:main-experimental-results}, the model trained with 150B is used for fair comparison, because not all models are finished training at the same iteration.
However, experiments in Section \ref{subsec:discussion-on-in-house-usage} use the model trained with 300B tokens, as \modelname{} Studio provided the 39B and 82B models trained with 300B tokens.

In our test loss from the encyclopedia corpus not included in \modelname{} corpus, we also observe the scaling law, as discovered in previous research \cite{brown2020language,kaplan2020scaling}.
Figure \ref{fig:scaling-law} in Appendix \ref{sec:scaling-law} shows that increasing model size and training longer give advantage. 

\subsection{Korean Tokenization}
\label{subsec:korean-tokenization}
Korean is an agglutinative language where noun is followed by particle and stem of verb or adjective is followed by endings, expressing various grammatical properties.
Properly tokenizing noun and particle, and stems and endings clarifies the semantics of each token. 
\citet{park2020empirical} introduce an empirical report that tokenization influences on performances of Korean LM. Overall, we need to design a sophisticated tokenization strategy suitable for Korean LM, different from its English counterpart.

We use morpheme-aware byte-level BPE as our tokenization method. GPT-2 and GPT-3 use byte-level BPE. However, unlike in English, non-English characters like `ㅎ', `하', or `한' are all split into three different unicode bytes. We alleviate the problem of byte-level BPE by applying morpheme analyzers. See Figure \ref{fig:mabbpe} in Appendix \ref{sec:movitation-of-tokenization} for motivation and detail.

We pre-split sentences by using space and morpheme obtained by an in-house morpheme analyzer. Our morpheme analyzer excludes most of non-Korean characters.
Using parts of the sentence pre-split by our morpheme analyzer, our morpheme-aware byte-level BPE learns the sentence in which most non-Korean characters are expressed as single byte characters.
We use HuggingFace's tokenizers library.\footnote{https://github.com/huggingface/tokenizers}

\begin{table*}[t!]
    \centering
    \small
    \begin{threeparttable}
    \begin{tabular}{lccccccc}
        \toprule
         & NSMC & \multicolumn{2}{c}{KorQuAD} & \multicolumn{2}{c}{AI Hub (BLEU)} & YNAT & KLUE-STS \\
         & (Acc)  & \multicolumn{2}{c}{(EM / F1)} & Ko$\rightarrow$En & En$\rightarrow$Ko & (F1) & (F1) \\
        \midrule
        Baseline & 89.66 & 74.04 & 86.66 & 40.34 & 40.41 & 82.64 & 75.93 \\
        \midrule
        137M & 73.11 & 8.87 & 23.92 & 0.80 & 2.78 & 29.01 & 59.54\\
        350M & 77.55 & 27.66 & 46.86 & 1.44 & 8.89 & 33.18 & 59.45 \\
        760M & 77.64 & 45.80 & 63.99 & 2.63 & 16.89 & 47.45 & 52.16 \\
        1.3B & 83.90 & 55.28 & 72.98 & 3.83 & 20.03 & 58.67 & 60.89 \\
        6.9B & 83.78 & 61.21 & 78.78 & 7.09 & 27.93 & 67.48 & 59.27 \\
        13B  & 87.86 & 66.04 & 82.12 & 7.91 & 27.82 & 67.85 & 60.00 \\
        39B  & 87.95 & 67.29 & 83.80 & 9.19 & 31.04 & 71.41 & 61.59 \\
        82B  & 88.16 & 69.27 & 84.85 & 10.37 & 31.83 & 72.66 & 65.14 \\
        \bottomrule
    \end{tabular}
    \end{threeparttable}
    \caption{Results of in-context few-shot tasks on sentiment analysis, question answering, machine translation, topic classification, and semantic similarity per model size. As baselines, we report the results of BERT-base for NSMC and KorQuAD, and Transformer for AI Hub from \citet{park2020empirical}. mBERT is used for YNAT and KLUE-STS from \citet{park2021klue}.}
    \label{table:in-context-main-result}
\end{table*}

\section{Experimental Results}
\label{sec:main-experimental-results}

\subsection{Experimental Setting}
We mainly use five datasets for evaluating in-context few-shot learning performance. Two of the five datasets come from KLUE \cite{park2021klue}, which is a massive benchmark of Korean NLU tasks and a work concurrent to our paper. We also use one additional in-house dataset for evaluating prompt-based optimization performance.

\noindent
\textbf{NSMC} is a movie review dataset from NAVER Movies.\footnote{https://github.com/e9t/nsmc} The task is binary sentiment classification, like SST-2 \cite{socher2013recursive}. It contains 150K of training data and 50K of test data. For few-shot experiments, we generate 12 sets, and each set consists of 70 examples randomly sampled from the training set. We average the test accuracies of 12 in-context 70-shot learning models. 

\noindent
\textbf{KorQuAD 1.0} \cite{lim2019korquad1} is a Korean version of machine reading comprehension dataset.\footnote{https://korquad.github.io/KorQuad\%201.0/}
It consists of 10,645 training passages with 66,181 training questions and 5,774 validation questions. The format of the dataset is similar to SQuAD 1.0 \cite{rajpurkar2016squad}. We follow the evaluation scheme of SQuAD v2.0 used in the work of \citet{brown2020language}, which uses test paragraph, corresponding four question-answer pairs, and test question as the input to GPT-3. 
In other words, our model is a zero-shot learner in the perspective of passage, but a four-shot learner in the perspective of question. We performed a single trial for each model size.

\noindent
\textbf{AI Hub Korean-English} corpus
consists of Korean-English parallel sentences from news, government websites, legal documents, etc.\footnote{https://aihub.or.kr/aidata/87} The corpus consists of 800K sentence pairs, and we randomly sample 1K pairs for evaluating on Ko $\rightarrow$ En and En $\rightarrow$ Ko translation tasks. 
We performed three random trials for each translation task. Our model is evaluated in four-shot learning and we use four different examples for each trial.
We use BLEU score for evaluation, where Moses and MeCab are used for comparison with the result of \citet{park2020empirical}.

\noindent
\textbf{YNAT} (Yonhap News Agency Topic Classification or KLUE-TC), one of the KLUE Benchmark tasks, is a topic classification problem with seven classes \cite{park2021klue}. It consists of 45K, 9K, and 9K annotated headlines for training, valid, and test sets, respectively. We average the test accuracies of 3 in-context 70-shot learners.

\noindent
\textbf{KLUE-STS}, another KLUE benchmark task, is a task to predict a sentence similarity between each pair of sentences, where the similarity score has a value between 0 and 5 \cite{park2021klue}. We use F1 score after binarizing the real-valued similarity as suggested in the KLUE paper. We average the test accuracies of 3 in-context 40-shot learners.

\noindent
\textbf{Query modification task} is a query modification task for AI speaker users.
The task targets the case where a single-turn FAQ system is already operating in AI Speakers. With the query that requires understanding of multi-turn information, the goal of the task is to convert the multi-turn query to a single-turn query, which can then be understood by a single-turn AI speaker. There are 1,326 test instances in total. See Appendix \ref{subsec:query-modification-task-appendix} for detail.

\noindent
\textbf{Baseline} 
We use the scores for baseline models, BERT and Transformer from \citet{park2020empirical}, and mBERT (BERT-Multilingual) from \citet{park2021klue}, for in-context learning experiments in Table \ref{table:in-context-main-result}, whereas mBERT \cite{devlin2019bert} and RoBERTa \cite{kang2020knowledge} are also used for p-tuning experiments in Table \ref{table:p-tuning}.

\subsection{In-context Few-shot Learning}
\label{subsec:in-context-few-shot-learning}

Table \ref{table:in-context-main-result} presents the results of few-shot learning on six tasks. In particular, we explore the performances of \modelname{} with mid-size parameters including 39B and 82B, which is not addressed in OpenAI GPT-3 paper~\cite{brown2020language} but can be more practical for real-world applications. 
Appendix \ref{subsec:nsmc-appendix} and \ref{subsec:ai-hub-translation-appendix} further explains more results of standard deviation and max performance of trials. 

Table \ref{table:in-context-main-result} shows that the performances of various in-context learning tasks monotonically increases as the model size increases. However, in-context learning ability of Ko$\rightarrow$En translation and KLUE-STS is much lower than baseline.
Especially for translation, we conjecture the poor performances on Ko$\rightarrow$En might result from lack of English ratio of our corpus. Also, more sophisticated prompt engineering might improve the results, which is future research direction.

\subsection{Prompt-based Tuning}
\label{subsec:prompt-based-tuning}

\begin{table}[t!]
    \centering
    \small
    \begin{threeparttable}
    \begin{tabular}{lc}
        \toprule
        Methods & Acc  \\
        \midrule
        Fine-tuning \\
        mBERT \cite{devlin2019bert} & 87.1\\
        ~~~w/ 70 data only & 57.2\\
        ~~~w/ 2K data only & 69.9\\
        ~~~w/ 4K data only & 78.0\\
        BERT \cite{park2020empirical} & 89.7\\
        RoBERTa \cite{kang2020knowledge} & 91.1\\
        \midrule
        Few-shot \\
        13B 70-shot & 87.9\\
        39B 70-shot & 88.0\\
        82B 70-shot & 88.2\\
        \midrule
        p-tuning \\
        137M w/ p-tuning & 87.2\\
        ~~~w/ 70 data only & 60.9\\
        ~~~w/ 2K data only & 77.9\\
        ~~~w/ 4K data only & 81.2\\
        13B w/ p-tuning & 91.7 \\
        ~~~w/ 2K data only & 89.5 \\
        ~~~w/ 4K data only & 90.7 \\
        ~~~w/ MLP-encoder & 90.3 \\
        39B w/ p-tuning & \textbf{93.0}\\
        \bottomrule
    \end{tabular}
    \end{threeparttable}
    \caption{Comparison results of p-tuning with fine-tuned LMs and in-context few-shot learning on NSMC. MLP-encoder means the result of replacing LSTM with MLP as the p-tuning encoder on 150K NSMC training data.}
    \label{table:p-tuning}
\end{table}

Table \ref{table:p-tuning} shows the results of prompt-based tuning (p-tuning)~\cite{liu2021gpt} on NSMC. Although in-context few-shot learning has already achieved near state-of-the-art performance on NSMC, p-tuning enables \modelname{} to outperform comparatives with no parameter update of the main model. It is worth noting that p-tuning with only 4K examples only provides comparable results to RoBERTa fine-tuned on 150K data. 
Considering the results in Table \ref{table:in-context-main-result} and Table \ref{table:nsmc-std-max} in Appendix \ref{subsec:nsmc-appendix}, we conjecture that p-tuning significantly enhances the robustness of \modelname{} as well as the accuracy. 

Furthermore, we explore the effects of p-tuning at the input side on performances for generation tasks with the experiments on our in-house query modification. As shown in Table \ref{table:p-tuning-query-modification}, p-tuning enables \modelname{} to consistently improve the input query qualities with a significant margin for both zero and three-shot scenarios. 
In larger models, the influence of the discrete prompt seems to be less. This result is similar to the trend discovered in \cite{lester2021power}, that as the scale of LM increases, competitive performance can be obtained even if the discrete prompt is not used at all.
To the best of our knowledge, this is the first report of applying input-side p-tuning to generation tasks with an in-context LM learner. 

These results also imply that when the backward gradients of GPT-3-scale model on input data are accessible, prompt optimization methods are feasible alternatives for enhancing representation power of large-scale LMs for NLP researchers and practitioners without large-scale GPU clusters.

\begin{table}[t!]
    \centering
    \small
    \begin{threeparttable}
    \begin{tabular}{llcc}
        \toprule
         Model sizes & Few-shots & p-tuning & BLEU  \\
        \midrule
        \multirow{4}{*}{13B} & \multirow{2}{*}{zero-shot} & $\times$ & 36.15 \\
        & & O & \textbf{58.04} \\
        \cline{2-4}
        & \multirow{2}{*}{3-shot} & $\times$ & 45.64\\
        & & O & \textbf{68.65} \\
        \midrule
        \multirow{4}{*}{39B} & \multirow{2}{*}{zero-shot} & $\times$ & 47.72 \\ 
        & & O & \textbf{73.80} \\
        \cline{2-4}
        & \multirow{2}{*}{3-shot} & $\times$ & 65.76\\
        & & O & \textbf{71.19} \\
        \bottomrule
    \end{tabular}
    \end{threeparttable}
    \caption{Results of p-tuning on in-house query modification task.}
    \label{table:p-tuning-query-modification}
\end{table}

\subsection{Effect of Tokenization}
\label{subsec:effect-of-tokenization}
We analyze the effects of morpheme-aware byte-level BPE, our tokenization method considering Korean linguistic characteristics. As baselines, we employ byte-level BPE and char-level BPE, two prevalent tokenization methods for pre-training LMs with English-centric corpora. 
It is noticeable that char-level BPE refers to the original BPE. It yields out-of-vocabulary (OOV), and some Korean character like `젝' is not included in char-level BPE tokens. The other two tokenization strategies do not make OOV tokens.
We use models of 1.3B parameters, which is a relatively small size, considering the heavy computation time of pre-training.
Nevertheless, it is enough to find evidence of tokenization effects. 

As shown in Table~\ref{table:MABBPE}, our method improves the performance of most tasks compared to the baselines. However, in Ko$\rightarrow$En task, morpheme analyzer makes the performance worse. On the other hand, char-level BPE makes much lower performance than byte-level BPE in YNAT. It is because that char-level BPE makes some OOV tokens, and some important words in a headline of YNAT data become hard to understand. For example, a character `젝' (jec) in a word `프로젝트' (project in English) is an OOV token in char-level BPE, which makes the test headline including `프로젝트' incomprehensive. Overall, it is worth noting that carefully designing language-specific tokenization is essential for training large-scale LMs for languages quite different from English in terms of their linguistic properties.

\begin{table*}[t!]
    \centering
    \small
    \begin{threeparttable}
    \begin{tabular}{lcccccc}
        \toprule
         & \multicolumn{2}{c}{KorQuAD} & \multicolumn{2}{c}{AI Hub (BLEU)} & YNAT & KLUE-STS\\
         &  \multicolumn{2}{c}{(EA / F1)} & Ko$\rightarrow$En & En$\rightarrow$Ko & (F1) & (F1)\\
        \midrule
        Ours & \textbf{55.28} & \textbf{72.98} & 3.83 & \textbf{20.03} & \textbf{58.67} & \textbf{60.89} \\
        \midrule
        byte-level BPE & 51.26 & 70.34 & \textbf{4.61} & 19.95 & 48.32 & 60.45 \\
        char-level BPE & 45.41 & 66.10 & 3.62 & 16.73 & 23.94 & 59.83 \\
        \bottomrule
    \end{tabular}
    \end{threeparttable}
    \caption{Effects of tokenization approaches on five tasks. \modelname{}-1.3B is used for evaluation.}
    \label{table:MABBPE}
\end{table*}

\section{Discussion on Industrial Impacts}

What change can large-scale LMs bring?
We claim ``accelerating the life-cycle of NLP ML operation'' as one of the possible answers. Unlike the protocol of most deep learning research where a model is trained with a well-collected dataset by ML experts and its corresponding well-defined objective function, there are several additional steps to make an AI product in a production-level pipeline, which yield tremendous communication overhead and costs. A platform with large-scale LMs may make huge progress by allowing only one non-developer, such as a service designer, to build the prototype system.

Section \ref{subsec:kogpt3-studio} introduces \modelname{} Studio as our distribution method of \modelname{}.
Section \ref{subsec:discussion-on-in-house-usage} introduces our three in-house usages of \modelname{} Studio.
Section \ref{subsec:opportunity} discusses possible extensions of \modelname{} Studio, prompt-based optimization, input module, and output module. 
Using the evidence above, Section \ref{subsec:no-code-ai-paradigm} discusses No/Low Code AI paradigm.

\subsection{\modelname{} Studio}
\label{subsec:kogpt3-studio}

\modelname{} Studio is the place for building and communicating the shared artifact generated by \modelname{}. 
\modelname{} Studio serves two functions, 1) it can provide a GUI interface, like the OpenAI Playground, and 2) support API end point in which the output can be easily acquired by an API call with diverse functions, including ones not yet provided by OpenAI Playground. These advanced functions are specified in Section \ref{subsec:opportunity}. 
Figure \ref{fig:studio1} in Appendix \ref{sec:details-on-studio} shows our GUI interface. 
The biggest advantage of \modelname{} Studio is that it allows rapid prototyping of AI-based services while minimizing the involvement of ML engineers.

\subsection{Case Studies on \modelname{} Studio}
\label{subsec:discussion-on-in-house-usage}

\begin{figure*}[t] 
\centering
\includegraphics[width=0.85\textwidth]{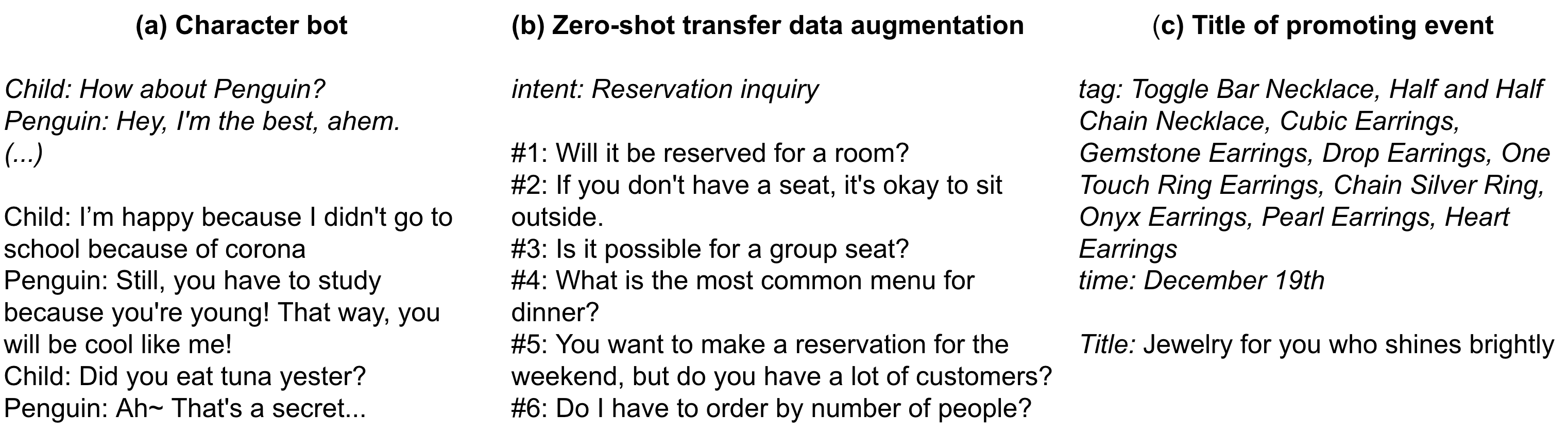}
\caption{Examples generated by \modelname{} with the prompts under three different tasks. Italic implies given prompts and non-italic corresponds to generated outputs. The examples are translated into English.}
\label{fig:discussion-examples}
\end{figure*}

This section shares three in-house applications powered by \modelname{} Studio, which are novel tasks with a large-scale LM as illustrated in Figure \ref{fig:discussion-examples}.
The three in-house usages share three properties below. First, it is non-trivial to define the objective function or to evaluate the models automatically. Second, the style of the inputs and outputs is easily controlled. Lastly, a product designer, without programming skill nor knowledge of AI, can easily make Proof-of-Concept (PoC) systems within few hours.

\subsubsection{Rapidly Prototyping Chatbots with Personalities}

This subsection discusses rapid prototyping of chatbots with personalities \cite{smestad2018chatbot} using \modelname{}. Our chatbot designers found that \modelname{} allows them to build a chatbot with the persona of a specific character using one or two lines of description on the character property and few dialog examples. This process can be used for producing many bots in metaverse applications. Figure \ref{fig:discussion-examples} (a) shows an example.

The style of the character can be controlled easily by changing a few dialog examples in the prompt. 
Knowledge in \modelname{} can also be implicitly extracted using the beginning of the prompt. For example, the knowledge of the famous can be reflected.  
Detailed discussion can be found in Appendix \ref{subsec:discussions-on-making-persona-chatbot}.

PoC can be easily available, and the following human-in-the-loop process can accelerate making a bot. Based on these functions, it is possible to quickly build a dialogue system of various characteristics. \modelname{} Studio also supports these functionalities.

\subsubsection{Zero-shot Transfer Data Augmentation}
\begin{table}[]
    \centering
    \footnotesize
    \setlength\tabcolsep{4pt}
    \begin{threeparttable}
    \begin{tabular}{lccccc}
        \toprule
        \multicolumn{6}{l}{Zero-shot (Acc)} \\
         & \multicolumn{5}{c}{\# of augmented samples ($k$)} \\
         $n$ & 5(1) & 10(2) & 15(3) & 25(5) & 125(30) \\
         \midrule
         0(0) & 60.8$_{9.3}$ & 68.9$_{4.0}$ & 71.9$_{2.7}$ & 74.8$_{2.5}$ & 78.0$_{2.3}$ \\
         \bottomrule\\
         \toprule
         \multicolumn{6}{l}{Few-shot (Acc)} \\
         & \multicolumn{5}{c}{\# of original samples ($n$)} \\
         $k$ & 1(1) & 2(1) & 3(1) & 4(1) & 5(1) \\
         \midrule
         0(0) & 26.8$_{6.0}$ & 52.0$_{4.9}$ & 64.7$_{5.2}$ & 76.5$_{4.4}$ & 83.0$_{3.0}$ \\
         25(5) & 79.2$_{2.5}$ & 81.2$_{2.5}$ & 82.6$_{2.6}$ & 83.4$_{1.9}$ & 84.3$_{2.0}$ \\
         125(30)  & 80.7$_{2.2}$ & 82.7$_{1.9}$ & 83.7$_{2.1}$ & 86.3$_{1.5}$ & 87.2$_{1.7}$\\
         \bottomrule
    \end{tabular}
    \end{threeparttable}
    \caption{Zero-shot and few-shot performances in zero-shot transfer data augmentation. $n$ denotes the number of original training (validation) instances per class, and $k$ denotes the number of generated instances for training (validation) per class. Subscripted values are standard deviation.}
    \label{table:zeroda}
\end{table}

The task is to build utterances tailored to user intent. Given the natural language name of the user's intent, corresponding utterances are generated. For example, if you give ``reservation query with one person'' as the user intent name, \modelname{} will output sentences like ``Is it OK for reservation with one person?'' 
We formulate this problem as in-context zero-shot transfer data augmentation. We give source-domain classes and corresponding examples for each source-domain class to the prompt. Source-domain classes are different from target-domain classes.

The name of intent can be simple, like ``reservation inquiry'' or complex, like ``Complaints about the degree of steak doneness''.
In in-house usages, a team for managing the quality of the product uses this function to make diverse utterances to validate the dialog system. 
The team reported that they could easily make diverse utterances of a intent with the complicated situation using \modelname{} Studio. 

We design a simple experiment to obtain quantitative results. We select 20 classes in an in-house intent corpus as the target domain and 6 classes with 5 examples each for the source domain.
Quantitative results using the 39B model are illustrated in Table \ref{table:zeroda}. See the details and discussions in Appendix \ref{subsec:zero-shot-transfer-data-augmentation-appendix}.

\subsubsection{Event Title Generation}
Event title generation is to generate the titles of an event for enhancing product advertisement in our e-commerce platforms. Similar to the significant effect of the product titles on CTR and revenue~\cite{zhang2019multi}, the product event title has a crucial influence on the product's success. Event title generation is formulated as a sequence-to-sequence task to transform keywords describing the product characteristics into an impressive event title. 

For achieving this, we ask an event designer to prepare five examples including event date and keywords as a prompt to \modelname{}. 
Within less than 10 minutes of designers' effort, \modelname{} Studio was able to generate the candidates of sales event titles with high quality.
Table \ref{table:advertisement} presents the quantitative results of the event title generation. We employ mT5-base (Multilingual T5) model~\cite{xue2020mt5} as a baseline. mT5-base has a size of 580M and is fine-tuned with 400K training data.
For human evaluation, we asked nine human experts to pick the best expression among the titles generated by GT, mT5, and \modelname{}.
As shown in Table~\ref{table:advertisement}, \modelname{} can yield high-quality titles comparable to GT. Interestingly, we find that higher BLEU scores with respect to GT do not guarantee higher qualities~\cite{mathur2020tangled}. On the contrary, it is worth noting that lower BLEU of \modelname{} implies that it can generate more creative titles, not using the exact words of GTs yet satisfying their qualities. Our system is also easy to control the theme that each designer wants to emphasize for the same keyword, such as discounting promotion, item brand, and product values. The detailed results are presented in Appendix \ref{subsec:headline-generation-appendix}. 

Unlike fine-tuned models, \modelname{} is easy to be adapted to the events of other domains by modifying the prompts. 
We also share usage of the advertisement headline task in the Appendix \ref{subsec:headline-generation-appendix}, where few training examples are available, but the prompt similar to the event title generation task achieves 99\% of appropriateness for the real service.  

\begin{table}[t!]
    \centering
    \footnotesize
    \begin{threeparttable}
    \begin{tabular}{lcccc}
        \toprule
        & BLEU & Win & Lose & Tie\\
        \midrule
        mT5 vs. GT & 13.28 & 0.311 & \bf{0.433} & 0.256\\
        \modelname{} vs. mT5  & - & \bf{0.456} & 0.350 & 0.194\\
        GT  vs. \modelname{} & 5.66 & 0.311  & 0.333 & \bf{0.356}\\
        
        \bottomrule
    \end{tabular}
    \end{threeparttable}
    \caption{Results of event title generation. GT denotes the ground truth title written by human experts. \textit{Win} means \textit{X} wins against \textit{Y} under \textit{X} vs. \textit{Y}. BLEU is the BLEU score of each model with its corresponding GT.}
    \label{table:advertisement}
\end{table}

\subsection{Opportunity of \modelname{} Studio}
\label{subsec:opportunity}
\modelname{} Studio can boost the ability of \modelname{} by multiple additional AI functions.
First, \textit{input gradient API}, which gives input gradient of \modelname{} can be applied to enhance the performance of local downstream tasks.
Even for the downstream task that the in-context learner performs well, prompt-based optimization can further boost the performance.
Section \ref{subsec:prompt-based-tuning} shows the possibility.
Our studio can be extended to supply input gradient function to support prompt-tuning in local machines.
Then each developer can also train their own prompt encoder using prompt-optimization methods, such as Autoprompt \cite{shin2020autoprompt}, p-tuning \cite{liu2021gpt}, or prompt tuning \cite{lester2021power}.

Second, \textit{prompt injection module} can be applied. \modelname{} can be used for an open-domain QA reader by using adequate documents retrieved by a retriever. In general, retrieving knowledge or similar examples can boost the performance of \modelname{}.

Finally, \textit{filters} for input and output are helpful for preventing misuse of \modelname{}. OpenAI API also provides a filter to monitor generations of sensitive or ethically inadequate sentences.

\subsection{No/Low Code AI Paradigm}
\label{subsec:no-code-ai-paradigm}


A typical machine learning development pipeline involves (1) problem definition and user research, (2) data gathering and annotation, (3) training and validating models, (4) deploying and operating machine learning systems (MLOps), (5) error analysis and user monitoring. It is an iterative process where any issue in one step propagates to other steps, and the need for revisiting the steps for revision and update constantly arises even after the model deployment.

This is especially tedious and resource-heavy, not only because this pipeline involves different expertise and different roles, but also because there is not a shared grounded artifact to facilitate the communication between the experts.

A single large-scale LM with GUI interfacing on a prompt, like \modelname{} Studio, can remarkably alleviate this problem. Specifically, the 2 $\sim$ 4th steps of the previous five processes can be combined into one step. In the unified phase, curating examples, prompt design, API parameter tuning, and API integration can take place at once.

It is notable that an approach with a single large-scale LM makes communication costs of experts be dramatically reduced. Through this, the prototype of desired AI product can be created within few hours.
Though many companies want to use AI technology, it is costly to make the companies and teams to use AI techniques and gather data for AI, Therefore, there have been several discussions about strategies for adopting AI technology \cite{raffel2020exploring}. An approach with a single large-scale LM provides a novel paradigm to research communities and industries.

No Code AI approach is powerful when fast iteration on PoC is beneficial or when services can be solely built with pure generation ability of large-scale model. Low Code AI approach can be used where it uses some training dataset \cite{liu2021makes} following by pre-processing code or input/output modules are required.

We discuss the challenges of achieving No/Low Code AI paradigm with large-scale LMs in Section \ref{sec:challenges} of the Appendix with detail.

\section{Conclusion}
We present HyperCLOVA, various billions-scale Korean-centric LMs. In particular, HyperCLOVA with 82B parameters shows state-of-the-art in-context zero-shot and few-shot performance and can further be boosted by prompt-based learning method.
We will share our model by \modelname{} Studio where non-developers can easily build their own AI-backed products.
We argue that a framework like \modelname{} Studio can potentially achieve No Code AI paradigm and hope that cases of such paradigm become popular, although opportunities and challenges coexist.

Our goal is to create an ecosystem using \modelname{} studio in Korea and help people not familiar with machine learning make their own AI models.

\section*{Acknowledgment}
The authors thank all the members of CLOVA, AI Lab, and Search of NAVER for devoted supporting and discussion. In particular, they thank Yonghwa Kim, Jin Hwan Suk, Jinsu Park, Hanah Lim, and the members of CLOVA ML X for qualitative evaluation. In addition, the authors thank NAVER Cloud for technically supporting the training environments of \modelname{}. Finally, the authors thank Reinald Kim Amplayo, Hwaran Lee, and Sohee Yang for proofreading.
\section*{Broader Impact Statement}
\label{sec:broader-impact-statement}

Since GPT3 was released, NLP and AI communities were impressed by the capability of its variants remarkably overwhelming the previous work. Despite their great success, these hyperscale pretrained LMs raise several severe concerning issues, which may harm the sustainability of AI and society.

\textbf{Misuse of large-scale LMs:} The case of Tay, the chatbot developed by Microsoft in 2016\footnote{https://bit.ly/3b6bL3o}, is one of the most well-known misusing examples. Recently, Luda, a Korean chatbot developed by a Korean startup, suffered from serious sexual abuse by malicious users\footnote{https://bit.ly/3tp1Rjs}. 
This situation brought a fundamental and social problem of whether AI can be an abused target to the surface. In Luda service, privacy issues were more critical from a legal perspective caused by incomplete data preprocessing for privacy-preserving. In addition to private information, hate speech data can lead to malicious misuse of language models when used as training data. Several GPT3 API applications also have reported these malicious usages and problematic generation results\footnote{https://www.wired.com/story/ai-fueled-dungeon-game-got-much-darker/}.

\textbf{Fairness, Bias, and Representation:} 
Another critical problem of Luda was biased and repulsive responses on various sensitive social values including gender and racism. Many studies have already reported that these biases from training data have significant influences on large-scale language models as well~\cite{abid2021persistent,garrido2021survey,shwartz2020neural}. To overcome these issues, many researchers argue the necessity of controllability when generating sentences such as filtering and investigate how to more effectively refine the data for debiasing~\cite{tamkin2021understanding}. 

\textbf{Excessive Energy Consumption:} 
Many researchers have serious concerns about too heavy energy consumption for training large-scale models, which have been recently reported by several analysis papers~\cite{patterson2021carbon,bender2021dangers}. Scaling raw presents more parameters and training data are essential for better performance, which inevitably makes the energy issue worse. A plausible alternative is to use energy-efficient hardware such as FPGA.  

\textbf{Efforts for Positive Directions:}
Despite all these concerns and side effects, large-scale LMs can provide significant and innovative benefits which cannot be expected from previous AI technologies. One of the most valuable functions of large-scale LMs is the possibility of No/Low Code AI. Despite many open-source AI libraries, developing AI systems and models with a certain level of quality still requires considerable effort, experience, and corresponding data, which are an entry barrier for AI democratization. However, No/Low Code AI allows industrial engineers and online service designers not familiar with machine learning to make a simple AI system or its prototypes rapidly. This contribution is a similar case to the success of office programs such as Microsoft office. We provided our \modelname{} Studio for our platform service designers, who showed surprising results and performances using our Studio with their creativity. The outputs and data generated by \modelname{} Studio are applied to our AI services. From this result, we found the possibility of No/Low Code AI with our \modelname{}, which is a meaningful step to realize AI democratization. Therefore, we need strong efforts to alleviate the problematic issues while benefiting from the values that large-scale LMs can provide.

\bibliography{custom}
\bibliographystyle{acl_natbib}
\clearpage
\appendix

\section{Details on Data}
\label{sec:appendix-details-on-data}

\subsection{Data Description}
\label{subsec:data-description}
As shown in Table \ref{table:table1}, 49\%, 15\%, and 13\% of the corpus come from blogs, community sites, and News corpus, respectively. 7\% of the corpus consists of comments from various websites mentioned above. 
5\% of the corpus comes from KiN\footnote{https://kin.naver.com/}, which is an online social QnA service similar to Quora. KiN corpus consists of open questions and answers written by users.
Note that our corpus also includes Korean Wikipedia, but the portion is very small (0.04\%). 
We also use Wikipedia for English and Japanese to enhance the ability of foreign languages.
Modu-corpus\footnote{https://corpus.korean.go.kr/} is a collection of various datasets collected by National Institute of Korean Language (NIKL). We use five datasets, including messenger, news, spoken language corpus, written language corpus, and web corpus from Modu-corpus. 
The data ratio per language is 97\%, 2\%, 0.5\%, 0.5\% in Korean, English, Japanese, and other languages, respectively.

\subsection{Data Cleaning}
\label{subsec:data-cleaning}
In a similar way to the work of \citet{brown2020language}, we train a logistic regression model that can measure the quality of each document. BERT feature of the document is used as an input. We assume high-quality encyclopedia documents as positive examples and crawled web documents as negative ones.
We exclude the documents predicted as low-quality.
To remove duplicated documents, we calculate the similarity of the documents with a hash function.
We also utilize an in-house spam filtering technique to remove undesired advertisements and documents.
Moreover, we exclude low-quality documents too short in length or too repetitive at levels of graphemes, numbers, or special characters.
In particular, we observe the review-type documents often contain too repetitive expressions because there is a policy on the length of writing a review.
Also, if the document contains too many swear words and slang, it is excluded.
Within the document, we remove duplicated sentences between title and content.
In the case of KiN corpus, if multiple answers are registered for one question, only the answers adopted by the questioner or the answers from certified experts, such as doctors or lawyers, were used.
Even if the answer was adopted, it was excluded if the author's reputation score was low.
We parse the HTML source code and use only meaningful parts of the HTML page for training the model.
For news-type documents, we remove typical parts that have insignificant information, such as the first line and the last phrase for affiliation.

\subsection{Data Anonymization}
\label{subsec:data-anonymization}
We mask the personal information such as resident registration number, email address, phone number, bank account number, credit card number, passport number, driver's license number, etc.
However, we remain non-critical parts of the numbers that can't be used to identify a person. For example, we extract the age and gender from resident registration number, location information from driver's license number, dialing code from a phone number, and domain address from email.

\subsection{Data Postprocessing}
\label{subsec:data-postprocessing}
For preprocessing, we also add prefix on the document like ``Web, Title: \$\{title\_name\}, Body: \$\{body\_text\}'', following the CTRL paper \cite{keskar2019ctrl}.

\section{Scaling Law}
\label{sec:scaling-law}

\begin{figure*}[t] 
\centering

\includegraphics[width=0.479\textwidth]{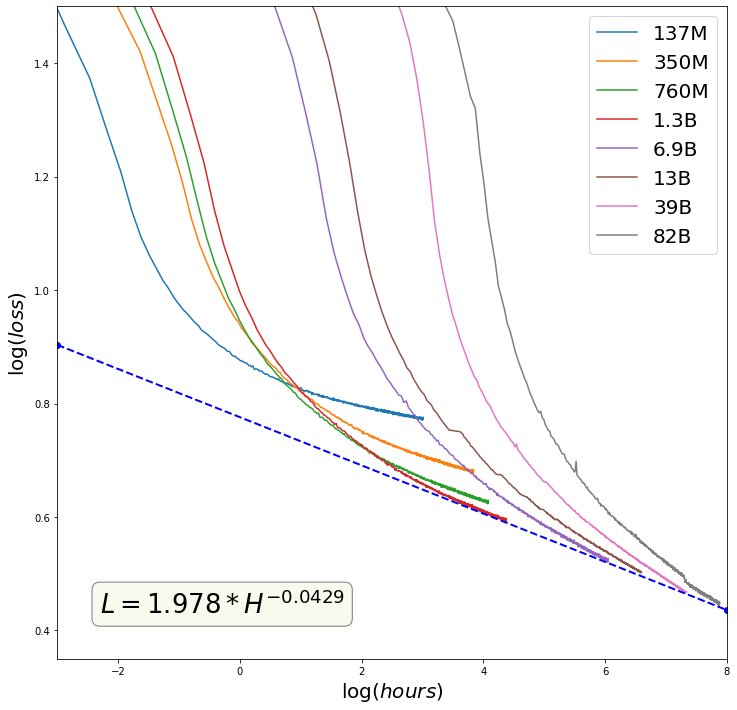}
\includegraphics[width=0.481\textwidth]{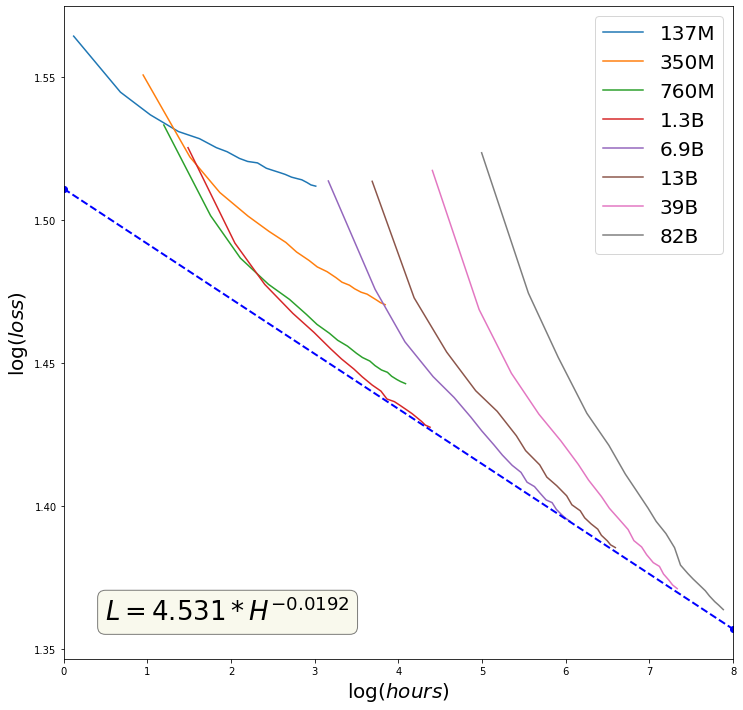}

\caption{Scaling law \cite{kaplan2020scaling,brown2020language} in training \modelname{} models with various parameters. The left figure presents the training and the right figure shows the loss on the testset of the Korean encyclopedia not contained in the training corpus.}
\label{fig:scaling-law}
\end{figure*}

Figure \ref{fig:scaling-law} shows the training and test loss patterns from a Korean encyclopedia corpus of \modelname{}. The results of \modelname{} are consistent to the scaling law pattern of GPT-3~\cite{kaplan2020scaling,brown2020language}.
\section{Details on Experiments}
\label{sec:details-on-main-experiements}

\subsection{NSMC}
\label{subsec:nsmc-appendix}

\begin{table}[t!]
    \centering
    \small 
    \begin{threeparttable}
    \begin{tabular}{lccc}
        \toprule
         & Mean & Std & Max \\
        \midrule
        137M & 73.11 & 3.11 & 77.19 \\
        350M & 77.55 & 4.68 & 82.93 \\
        760M & 77.64 & 7.25 & 85.03 \\
        1.3B & 83.90 & 1.90 & 86.03 \\
        6.9B & 83.78 & 2.76 & 87.62 \\
        13B & 87.86 & 0.52 & 88.79 \\
        39B & 87.95 & 0.54 & 88.87 \\
        82B & 88.16 & 0.75 & 89.16 \\
        \bottomrule
    \end{tabular}
    \end{threeparttable}
    \caption{Mean, standard derivation, and max accuracy on NSMC.}
    \label{table:nsmc-std-max}
\end{table}

Table  \ref{table:nsmc-std-max} shows the statistics on the performance of \modelname{} in NSMC.

\subsection{AI Hub Translation}
\label{subsec:ai-hub-translation-appendix}

\begin{table}[t!]
    \centering
    \small 
    \begin{threeparttable}
    \begin{tabular}{lcccccc}
        \toprule
         & \multicolumn{3}{c}{Ko$\rightarrow$En} & \multicolumn{3}{c}{En$\rightarrow$Ko} \\
         & Mean & Std & Max & Mean & Std & Max \\
        \midrule
        137M & 0.80 & 0.06 & 0.87 & 2.78 & 0.35 & 3.09 \\
        350M & 1.44 & 0.06 & 1.52 & 8.89 & 0.37 & 9.24 \\
        760M & 2.63 & 0.10 & 2.75 & 16.89 & 0.83 & 17.80 \\
        1.3B & 3.83 & 0.08 & 3.92 & 20.03 & 0.13 & 20.15 \\
        6.9B & 7.09 & 0.20 & 7.29 & 27.93 & 0.58 & 28.42 \\
        13B & 7.91 & 0.33 & 8.19 & 27.82 & 0.25 & 28.10 \\
        39B & 9.19 & 0.26 & 9.49 & 31.04 & 0.78 & 31.93 \\
        82B & 10.37 & 0.22 & 10.53 & 31.83 & 0.70 & 32.63 \\
        \bottomrule
    \end{tabular}
    \end{threeparttable}
    \caption{Mean, standard derivation, and max accuracy on AI Hub translation tasks.}
    \label{table:translation-std-max}
\end{table}

Table \ref{table:translation-std-max} shows the statistics on the performance of \modelname{} in AI Hub translation tasks.

\subsection{Query Modification Task}
\label{subsec:query-modification-task-appendix}

Table \ref{table:query-modification-example} and Table \ref{table:query-modification-prompt} show the example and the prompt for the query modification task.

\begin{table}[t!]
    \centering
    \small
    \begin{tabular}{l}
        \hline
        \hline
        Example 1: \\
        \hline
        사용자: 아이유 노래 틀어줘\\ 
        (User: Play IU's track)\\         
        \hline
        스피커: 노래를 재생합니다. \\
        (AI Speaker: I am playing the track.) \\
        \hline
        사용자: 몇 살이야 \\
        (User: How old?)\\         
        \hline
        \textbf{사용자의 최종 의도: 아이유 몇 살이야} \\
        \textbf{(Modified query: How old is IU?)}\\
        \hline
        \hline
        Example 2: \\
        \hline
        사용자: 비행기는 누가 만들었어 \\
        (User: Who invented airplane?)\\ 
        \hline
        스피커: 라이트형제요. \\ 
        (AI Speaker: Wright brothers did.)\\ 
        \hline
        사용자: 동생 이름 뭐야 \\
        (User: What is the younger's name?.) \\
        \hline
        \textbf{사용자의 최종 의도: 라이트 형제 동생 이름 뭐야?} \\
        \textbf{(Modified query: What is the younger one's name of} \\\textbf{Wright brothers?)}\\
        \hline\hline
    \end{tabular}
    \caption{Examples of user query modified by \modelname{}. English sentences are translated by a human expert.}
    \label{table:query-modification-example}
\end{table}

\begin{table}[t!]
    \centering
    \small
    \begin{tabular}{l}
        \hline
        [P][P][P][P][P][P][P][P][P][P]\\
        \\
        \# 예제1\\
        (\# Example 1)\\
        사용자: 아이유 앨범 뭐있어\\
        (User: What are the names of some albums of IU?)\\
        스피커: 아이유의 대표 앨범으로는 Love poem, Palette, \\
        \ \ \ \ \ \ \ \ \ \ \ \ \ CHAT-SHIRE가 있어요.\\
        (AI Speaker: IU's signiture albums include Love poem, \\
        \ \ \ \ \ \ \ \ \ \ \ \ \ Palette, and CHAT-SHIRE.)\\
        사용자: 가장 신나는 앨범이 뭐야\\
        (User: Which one is the most exciting album?)\\
        --\\
        $ $[P][P][P] 사용자의 [P][P] 의도: Love poem, Palette, \\
        \ \ \ \ \ \ \ \ \ \ \ \ \ CHAT-SHIRE 중 가장 신나는 앨범이 뭐야\\
        ($ $[P][P][P] User's [P][P] intent: Among Love poem,  \\
        \ \ \ \ \ \ \ \ \ \ \ \ \ Palette, and CHAT-SHIRE, which one is the\\
        \ \ \ \ \ \ \ \ \ \ \ \ \ most exciting album?)\\
        
        \\
        \# 예제2\\
        (\# Example 2)\\
        사용자: 평창 동계올림픽은 몇년에 했어?\\
        (User: When did the PyeongChang Olympics take place?)\\
        스피커: 2018년입니다.\\
        (AI Speaker: It is 2018.)\\
        사용자: 그때 미국 대통령이 누구야\\
        (User: Who was the president of the United States at that\\
        \ \ \ \ \ \ \ \ \ \ \ \ \ time?)\\

        --\\
        $ $[P][P][P] 사용자의 [P][P] 의도: 2018년 미국 대통령이\\
        \ \ \ \ \ \ \ \ \ \ \ \ \ 누구야\\
        ($ $[P][P][P] User's [P][P] intent: Who was the president of \\
        \ \ \ \ \ \ \ \ \ \ \ \ \ US in 2018?)\\
        \\
        \# 예제3\\
        (Example 3)\\
        사용자: 삼성전자 주가 얼마야\\
        (User: What is Samsung Electronics' share price?)\\
        스피커: 8만2천원입니다.\\
        (AI Speaker: It is 82,000 Won.)\\
        사용자: LG전자는\\
        (User: How about LG Electronics?)\\
        --\\
        $ $[P][P][P] 사용자의 [P][P] 의도: LG전자 주가 얼마야\\
        ($ $[P][P][P] User's [P][P] intent: What is LG Electronics' \\
        \ \ \ \ \ \ \ \ \ \ \ \ \ share price?)\\
        \\
        \# 예제4\\
        (Example 4)\\
        \hline
    \end{tabular}
    \caption{Used prompts of query modification task. [P] denotes a token for continuous prompt.}
    \label{table:query-modification-prompt}
\end{table}

\subsection{Discussions on Making Persona Chatbot}
\label{subsec:discussions-on-making-persona-chatbot}

Recent chit-chat with the neural model, like Meena and Blender, shows impressive conversational performance \cite{Humeau2020Poly-encoders:,adiwardana2020towards,roller2020recipes}. However, such a conversation system uses a lot of data, and it cannot make a new style of conversational system in an instant. There are also plenty of researches on style transfer. However, these methods do not control the detailed style of the conversational system \cite{smith2020controlling}.

There also exist some hallucination issues. Retrieved knowledge can alleviate this problem \cite{shuster2021retrieval}. A pre-trained reader can also get advantages if the pre-trained LM itself also performs well for open-domain QA, as shown in T5 and FiD in open-domain question answering \cite{raffel2020exploring,izacard2021leveraging}.

\subsection{Zero-shot Transfer Data Augmentation}
\label{subsec:zero-shot-transfer-data-augmentation-appendix}

\modelname{} does not always make sentences which is fit to the target intent class. However, even when people simply fill in the utterances that fit their intent, it is difficult to create various patterns, and data collectors struggle to make many utterances because of this problem. Data collectors can easily make a corpus by selecting sentence candidates created by \modelname{}. 
Our corpus designer also found that generating dialect or converting standard language to dialect is also easily available, showing the capability of data augmentation with \modelname{}. 

Note that this experiment is zero-shot transfer data augmentation, and examples of a different class from target classes are used as in-context examples. We use a total of 30 examples from six source classes and randomly sample three source classes and corresponding 15 examples to put into the prompt. 
For classification, an in-house BERT-based model is used.

In our experiment, sentences for 18 classes are generated well (like 80\% $\sim$ 90\%), whereas sentences for 2 classes are not generated well (like 10\% $\sim$ 20\%). 

Similar concurrent works are conducted from \citet{schick2021generating}. However, their study can only be applicable for NLI, which is a well-defined task, has good datasets, and has pre-trained models for the task.

\begin{table}[t!]
    \centering
    \small
    \begin{tabular}{l}
        \hline
        사용자 인텐트에 맞는 문장 5개를 만드시오.\\
        (Create five sentences which match the user intent.)\\
        \\
        @ 사용자인텐트 : 포장 가능 문의\\
        (@ User intent: Inquiry on takeout)\\
        예시 발화\\
        (Example utterances)\\
        1. 칼국수나 돈까스 같은 음식도 포장되요?\\
        (1. Can I get food like Kalguksu or pork cutlet to go?)\\
        2. 죄송한데 테이크아웃 되죠?\\
        (2. Excuse me, can I takeout this?)\\
        3. 메뉴 포장 되나요?\\
        (3. Can I get this menu to go?)\\ 
        4. 아이스크림 포장해주세요\\
        (4. I'd like to get this ice cream to go.)\\
        5. 집에서도 먹을 수 있게 포장해주시나요?\\
        (5. Can I get this menu to go so I can eat this at home?)\\
        \\
        @ 사용자인텐트 : 배달음식 환불\\
        (@ User intent: refund on delivery food)\\
        예시 발화\\
        (Example utterances)\\
        1. 보쌈에서 시큼한 냄새가 나는데 환불부탁드립니다\\
        (1. Bossam smells sour, please give me a refund.)\\
        2. 메뉴가 잘못 배달 되었습니다. 환불부탁드립니다\\
        (2. The menu was delivered incorrectly. Please give me a)\\
        \ \ \ \ refund.)\\
        3. 간장게장 맛이 이상해요. 환불 가능 한가요?\\
        (3. Soy Sauce Marinated Crab tastes weird. Can I get a\\
        \ \ \ \ refund?)\\
        4. 치킨이 너무 식어서 왔어요. 환불 부탁드려요\\
        (4. The chicken is too cold. I'd like a refund, please.)\\
        5. 음식에서 벌레가 나왔네요. 환불 해주세요\\
        (5. There's a bug in the food. Please give me a refund.)\\
        \\
        @ 사용자인텐트 : 예약 좌석 문의\\
        (@ User intent: Inquiry on seat reservation)\\
        예시 발화\\
        (Example utterances)\\
        1. 20명이 가려는데, 자리가 충분한가요?\\
        (1. There are 20 people going, is there enough room?)\\
        2. 조용히 식사하고 싶은데 조용한 자리가 있을까요?\\
        (2. I'd like to have a quite meal, is there a quiet table?)\\
        3. 유모차를 가지고 들어가서 아이 눕혀놓고 싶은데 \\
        \ \ \ \ 마땅한 자리가 있을까요?\\
        (I'd like to take the stroller in and put the child down, is\\
        \ \ \ \ there a suitable seat?)\\
        4. 15명정도 간단하게 가족 친척들과 아기돌잔치 할만한 \\
        \ \ \ \ 자리있을까요?\\
        (4. Is there a place for 15 people to have a first birthday)\\
        \ \ \ \ party with their family and relatives?)\\
        5. 분리된 예약석이 있을까요?\\
        (5. Is there a separate reserved seat?)\\
        \\
        @ 사용자인텐트 : \{\textit{intent-name}\}\\
        (@ User intent:\{\textit{intent-name}\})\\ 
        예시 발화\\
        (Example utterances)\\
        1.\\
        \hline
    \end{tabular}
    \caption{A prompt for zero-shot transfer data augmentation.}
    \label{table:data-augmentation-prompt}
\end{table}

\subsection{Advertisement Design}
\label{subsec:headline-generation-appendix}

Table \ref{table:advertisement-design-prompt} and \ref{table:advertisement-design-control-prompt} show the example prompt for the event title generation task.
Table \ref{table:advertisement-design-comparison-example} shows a qualitative comparison between mT5 and our model.

Similar to the event title generation task, the product designer also does the advertisement headline generation task in a similar way. In this task, there is no training data which could be used due to data privacy issue. Nevertheless, \modelname{} with a similar style of event title generation task successfully generates an advertisement headline. 

Table \ref{table:headline-prompt} shows the prompt.
Three different prompts are used for advertisement headline generation, and the generated sentence which is most similar to the product name, which is an input of the task, is selected.
A similarity score is calculated by the cosine similarity score using a feature of the in-house BERT.
The product designer evaluates that 99\% of generated sentences are appropriate for the real service.

\begin{table}[t!]
    \centering
    \small
    \begin{tabular}{l}
        \hline
        태그: 댕댕이옷, 크리스마스, 따뜻한강아지옷,\\
        \ \ \ \ \ \ \ \ \ \ 강아지코스튬\\
        (Tags: daengdaeng-i\footnote{Korean neologism meaning puppy} clothes, Christmas, warm puppy \\
        \ \ \ \ \ \ \ \ \ \ clothes, puppy costume) \\
        날짜: 12월 23일\\
        (Date: December 23) \\
        제목: 겨울시즌 댕댕이를 위해\\
        (Title: For puppies in the winter season) \\
        \#\#\#\\
        태그: 암막커튼, 침실커튼, 방한커튼, 인테리어커튼, \\
        \ \ \ \ \ \ \ \ \ 아이방커튼, 거실커튼, 방풍커튼, \\
        \ \ \ \ \ \ \ \ \ 가리개커튼, 작은창커튼\\
        (Tags: blackout curtains, bedroom curtains, cold protection  \\ 
        \ \ \ \ \ \ \ \ \ \ curtains, interior curtains, children's room curtains, \\
        \ \ \ \ \ \ \ \ \ \ living room curtains, windproof curtains, blind \\
        \ \ \ \ \ \ \ \ \ \ curtains, small window curtains) \\
        날짜: 11월 1일\\
        (Date: November 1) \\
        제목: 찬바람 막는 방한커튼\\
        (Title: Cold protection curtains for blocking cold winds) \\
        \#\#\#\\
        태그: 명품구두, 여자들의로망, 여름구두\\
        (Tags: luxury shoes, women's dreams, summer shoes) \\
        날짜: 7월7일\\
        (Date: July 7) \\
        제목: 보기만 해도 행복한 명품 슈즈\\
        (Title: Luxury shoes that make you happy  \\
        \ \ \ \ \ \ \ \ \ \ just by looking at them) \\
        \#\#\#\\
        태그: 유아동복, 주니어쇼핑몰, 아동원피스, \\
        \ \ \ \ \ 아동맨투맨, 아동바지, 아동레깅스, 아동모자, \\
        \ \ \ \ \ 아동가방,아동양말, 아동신발\\
        (Tags: children's clothes, junior shopping mall, children's  \\ 
        \ \ \ \ \ \ \ \ \ \ dresses,  children's sweatshirts, children's pants,  \\
        \ \ \ \ \ \ \ \ \ \ children's leggings, children's hats, children's bags, \\
        \ \ \ \ \ \ \ \ \ \ children's socks, children's shoes) \\
        날짜: 2월 26일\\
        (Date: February 26) \\
        제목: 주목받는 신학기 코디제안\\
        (Title: New semester style suggestion for attracting\\
        \ \ \ \ \ \ \ \ \ \ attention) \\
        \#\#\#\\
        태그: 스마트워치, 스마트밴드, 웨어러블디바이스\\
        (Tags: Smart watch, smart band, wearable device) \\
        날짜: 12월 7일\\
        (Date: December 7) \\
        제목: 시계도 스마트하게 사용해봐\\
        (Title: Try to use your watch smartly) \\
        \#\#\#\\
        태그: 커피머신, 에스프레소머신, 커피메이커\\
        (Tags: coffee machine, espresso machine, coffee maker) \\
        날짜: 4월 13일\\
        (Date: April 13) \\
        제목: \textit{커피 한 잔의 여유를 위한 커피머신}\\
        (Title: \textit{Coffee machine for relaxing with a cup of coffee}) \\
        \hline
    \end{tabular}
    \caption{Prompt for product event title generation.}
    \label{table:advertisement-design-prompt}
\end{table}

\begin{table}[t!]
    \centering
    \small
    \begin{tabular}{l}
        \hline
        상품명: 디퓨저꽃 디퓨져스틱 방향제 리드스틱 머스타드\\
        \ \ \ \ 7종\\
        날짜: 2021년 3월 29일\\
        카테고리: 기타아로마/캔들용품\\
        브랜드: 캔들날다 메이릴리\\
        태그: 72993$\wedge$방향제만들기|64225$\wedge$디퓨저diy|189638$\wedge$\\
        \ \ \ \ 디퓨저리드|139746$\wedge$디퓨저만들기|198335$\wedge$\\
        \ \ \ \ 디퓨저만들기재료|379365$\wedge$인테리어디퓨저\\
        속성: |\\
        광고문구: 봄을 부르는 향기\\
        \#\#\#\#\#\#\\
        상품명: LYNN 린 차이나 프릴 블라우스\\
        날짜: 2021년 3월 29일\\
        카테고리: 블라우스/셔츠\\
        브랜드: 린\\
        태그: |\\
        속성: 핏$\wedge$기본핏|패턴$\wedge$무지|디테일$\wedge$프릴/러플|총기장$\wedge$\\
        \ \ \ \ 기본/하프|주요소재$\wedge$폴리에스테르|소매기장$\wedge$반팔\\
        광고문구: 여성스러운 프릴 블라우스\\
        \#\#\#\#\#\#\\
        상품명: 맥 아이 섀도우 1.5g\\
        날짜: 2021년 3월 29일\\
        카테고리: 아이섀도\\
        브랜드: 맥\\
        태그: 75984$\wedge$선물로좋은|76503$\wedge$포인트주기좋은|281615\\
        \ \ \ \ $\wedge$자연스러운발색|240838$\wedge$지속력좋은|235326$\wedge$\\
        \ \ \ \ 포인트연출|665375$\wedge$파우더리|1228492$\wedge$\\
        \ \ \ \ 부드러운사용감|836046$\wedge$자연스러운스모키|5279$\wedge$\\
        \ \ \ \ 청순메이크업|78091$\wedge$선물포장\\
        속성: 형태$\wedge$압축/팩트형|세부제품특징$\wedge$고운입자|\\
        \ \ \ \ 세부제품특징$\wedge$은은함|세부제품특징$\wedge$웜톤용|색상$\wedge$\\ 
        \ \ \ \ 골드|주요제품특징$\wedge$고발색|색상$\wedge$핑크|세부제품특징$\wedge$\\
        \ \ \ \ 눈매연출|세부제품특징$\wedge$펄있음|주요제품특징$\wedge$\\
        \ \ \ \ 부드러운발림|색상$\wedge$브라운|타입$\wedge$싱글|주요제품특징$\wedge$\\
        \ \ \ \ 지속력\\
광고문구: 매트한 질감과 선명한 발색\\
\#\#\#\#\#\#\\
상품명: 케이스 아쿠아텍스 이지클린 패브릭 원단 저상형\\
\ \ \ \ 패밀리 침대 SS,Q\\
날짜: 2021년 05월 17일\\
카테고리: 패밀리침대\\
브랜드: ss퍼니처\\
태그: 사이즈$\wedge$슈퍼싱글+퀸|부가기능$\wedge$안전가드포함|\\
\ \ \ \ 프레임$\wedge$저상형|자재등급$\wedge$E0(친환경)|\\
\ \ \ \ 부가기능$\wedge$유해물질차단|프레임소재$\wedge$패브릭\\
속성: 5554855641\\
광고문구: \textit{안전한 소재로 제작된 저상형 패밀리 침대}\\
        \hline
    \end{tabular}
    \caption{Prompt for advertisement headline design.}
    \label{table:headline-prompt}
\end{table}

\begin{table}[t!]
    \centering
    \small
    \begin{tabular}{l}
        \hline
        Product name: Diffuser flower diffuser stick air\\
        \ \ \ \ freshener reed stick mustard 7 kinds\\
        Date: March 29, 2021\\
        Category: other aroma/candle supplies\\
        Brand: Candlenalda maylily\\
        Tag: 72993$\wedge$making air freshener|64225$\wedge$diffuser diy|\\
        \ \ \ \ 189638$\wedge$diffuser reed|139746$\wedge$making diffuser|\\
        \ \ \ \ 198335$\wedge$diffuser making material|\\
        \ \ \ \ 379365$\wedge$interior diffuser\\
        Attribute: |\\
        Ads. headline: The scent of calling spring\\
        \#\#\#\#\#\#\\
        Product name: LYNN lynn china frill blouse\\
        Date: March 29, 2021\\
        Category: blouse/shirt\\
        Brand: Lynn\\
        Tag: |\\
        Attribute: fit$\wedge$basic fit|pattern$\wedge$plain|details$\wedge$frill/ruffle|\\
        \ \ \ \ legnth$\wedge$basic/half|material$\wedge$Polyester|sleeve$\wedge$short\\
        Ads. headline: Feminine frilled blouse\\
        \#\#\#\#\#\#\\
        Product name: Mac eye shadow 1.5g\\
        Date: March 29, 2021\\
        Category: eye shadow\\
        Brand: Mac\\
        Tag: 75984$\wedge$good for gifts|76503$\wedge$good for points|\\
        \ \ \ \ 281615$\wedge$natural color|240838$\wedge$long lasting|\\
        \ \ \ \ 235326$\wedge$point makeup|665375$\wedge$foundary|\\
        \ \ \ \ 1228492$\wedge$soft feeling|836046$\wedge$natural smokey|\\
        \ \ \ \ 5279$\wedge$pure makeup|78091$\wedge$gift wrapping\\
        Attribute: form$\wedge$compressed/fact-type|detailed features\\
        \ \ \ \ $\wedge$fine particles|detailed features$\wedge$subtlety|\\ 
        \ \ \ \ detailed features$\wedge$warm tone|color$\wedge$gold|\\
        \ \ \ \ main features$\wedge$high color|color$\wedge$pink|detailed\\
        \ \ \ \ features$\wedge$eye makeup|detailed features$\wedge$pearl|\\
        \ \ \ \ main features$\wedge$soft application|color$\wedge$brown|\\
        \ \ \ \ type$\wedge$single|main features$\wedge$long lasting\\
        Ads. headline: Matte texture and vivid color\\
        \#\#\#\#\#\#\\
        Product name: Case aquatex easy-clean fabric\\
        \ \ \ \ low-rise family bed SS,Q\\
        Date: May 17, 2021\\
        Category: family bed\\
        Brand: ssfurniture\\
        Tag: size$\wedge$super single+queen|Additional function\\
        \ \ \ \ $\wedge$include a safe guard|frame$\wedge$low-rise|\\
        \ \ \ \ material grade$\wedge$E0(eco)|additional function$\wedge$\\
        \ \ \ \ block harmful substances|frame material$\wedge$fabric\\
        Attribute: 5554855641\\
        Ads. headline: \textit{Low-rise family bed made of safe materials}\\
        \hline
    \end{tabular}
    \caption{English translation for Table~\ref{table:headline-prompt}.}
    \label{table:headline-prompt-eng}
\end{table}

\begin{table}[t!]
    \centering
    \small
    \begin{tabular}{ll}
        \toprule
        Models & Product event titles \\
        \hline
        \multirow{2}{*}{mT5} & 봄맞이 인테리어 발매트 모음전 \\
        & (Interior foot-mat event for spring season.)\\ 
        \multirow{2}{*}{\modelname{}} & 욕실 분위기를 바꿔줄 아이템\\
        & (Items that can change bathroom mood.)\\
        \hline
        \multirow{2}{*}{mT5} & 타이니러브 바운서 \\
        & (Tiny love bouncer.) \\
        \multirow{2}{*}{\modelname{}} & 엄마와 아기를 위한 편안함\\
        & (Comfort for mommy and baby.) \\
        \hline
        \multirow{2}{*}{mT5} & 한끼 요리 탕요리 반조리 \\
        & (A meal, stew, semi-cooked.) \\
        \multirow{2}{*}{\modelname{}} & 저녁 걱정 뚝! 간편한 탕요리\\
        & (No worry on dinner! \\
        & \ Simple semi-cooked stew.) \\
        \hline
        \multirow{2}{*}{mT5} & 가을맞이 면접룩 기획전 \\
        & (Interview fashion event for fall season.) \\
        \multirow{2}{*}{\modelname{}} & 면접 때 입을 옷 고민하지 마세요\\
        & (No worry on your fashion \\
        & \ for the interview.) \\
        \bottomrule
    \end{tabular}
    \caption{Examples of product event titles generated by mT5 and \modelname{}. English phrases in parenthesis are translated by human experts for preserving their nuances and semantics. }
    \label{table:advertisement-design-comparison-example}
\end{table}

\begin{table}[t!]
    \centering
    \small
    \begin{tabular}{l}
        \hline
        키워드: 캔디주얼리, 프로포즈목걸이, 커플링, 은반지,\\ 
        \ \ \ \ 다이아가드링, 로즈골드목걸이, 하트귀걸이, \\
        \ \ \ \ 하트목걸이\\
        (Keywords: candy jewelry, proposal necklace, coupling, \\ 
        \ \ \ \ \ silver ring, diamond guard ring, rose gold necklace, \\
        \ \ \ \ \ heart earring, heart necklace \\
        날짜: 2021년3월7일\\
        (Date: March 7, 2021) \\
        제목: 화이트데이 커플주얼리 세일\\
        (Title: White Day Couple Jewelry Sale) \\
        \\
        키워드: 수입그릇, 빈티지그릇, 법랑냄비, 수저세트,\\
        \ \ \ \ 튼튼한컵, 레트로냄비\\
        (Keywords: imported bowl, vintage bowl, enamel pot,  \\ 
        \ \ \ \ \ spoon and chopsticks set, strong cup, retro pot \\
        날짜: 2020년4월21일\\
        (Date: April 21, 2020) \\
        제목: 주방용품 해외직구 할인전\\
        (Title: Kitchenware overseas direct purchase \\ 
        \ \ \ \ \ discount exhibition) \\
        \\
        키워드: 미세먼지, 차량용핸드폰거치대, 세차용품,\\
        \ \ \ \ 자동차용품, 차량용품, 차량무선충전거치대,\\
        \ \ \ \ 차량악세사리, 논슬립패드, 자동차악세사리\\
        (Keywords: fine dust, mobile phone holder for vehicles, \\ 
        \ \ \ \ \ car washing products, automobile supplies, vehicle \\
        \ \ \ \ \ supplies, vehicle wireless charging cradle \\
        \ \ \ \ \ vehicle accessories, non-slip pads, car accessories) \\
        날짜: 2021년4월1일\\
        (Date: April 1, 2021) \\
        제목: 각종 차량용품 할인 모음전\\
        (Title: Collection of discounts on various vehicle supplies) \\
        \\
        키워드: 슬리퍼, 실내용슬리퍼, 사무용슬리퍼, 하이힐,\\
        \ \ \ \ 봄신상신발, 봄신발, 여자슬리퍼, 여성슬리퍼,\\
        \ \ \ \ 여성하이힐, 여자하이힐\\
        (Keywords: slippers, indoor slippers, office slippers, high \\ 
        \ \ \ \ \ heels, spring new arrival shoes, spring shoes, women's \\
        \ \ \ \ \ slippers, female slippers, women's high heels, \\
        \ \ \ \ \ female high heels) \\
        날짜: 2021년3월1일\\
        (Date: March 1, 2021) \\
        제목: 봄 여성 사무용 슬리퍼 하이힐 SALE\\
        (Title: Spring women's office slippers high heels SALE) \\
        \\
        키워드: 봄신상, 명품악세사리, 링귀걸이, 꽃머리끈,\\
        \ \ \ \ 명품키링, 머리끈, 악세사리\\
        (Keywords: spring new arrival, luxury accessories, ring  \\ 
        \ \ \ \ \ earrings, flower headbands, luxury key ring, headband, \\
        \ \ \ \ \ accessories) \\
        날짜: 2020년8월8일\\
        (Date: August 8, 2020) \\
        제목: \textit{악세서리 인기제품 할인전}\\
        (Title: \textit{Accessory popular products discount exhibition}) \\
        \hline
    \end{tabular}
    \caption{Controlling style by change in-context examples for product event title generation.}
    \label{table:advertisement-design-control-prompt}
\end{table}

\section{Details on Studio}
\label{sec:details-on-studio}

\begin{figure*}[t] 
\centering
\includegraphics[width=1.00\textwidth]{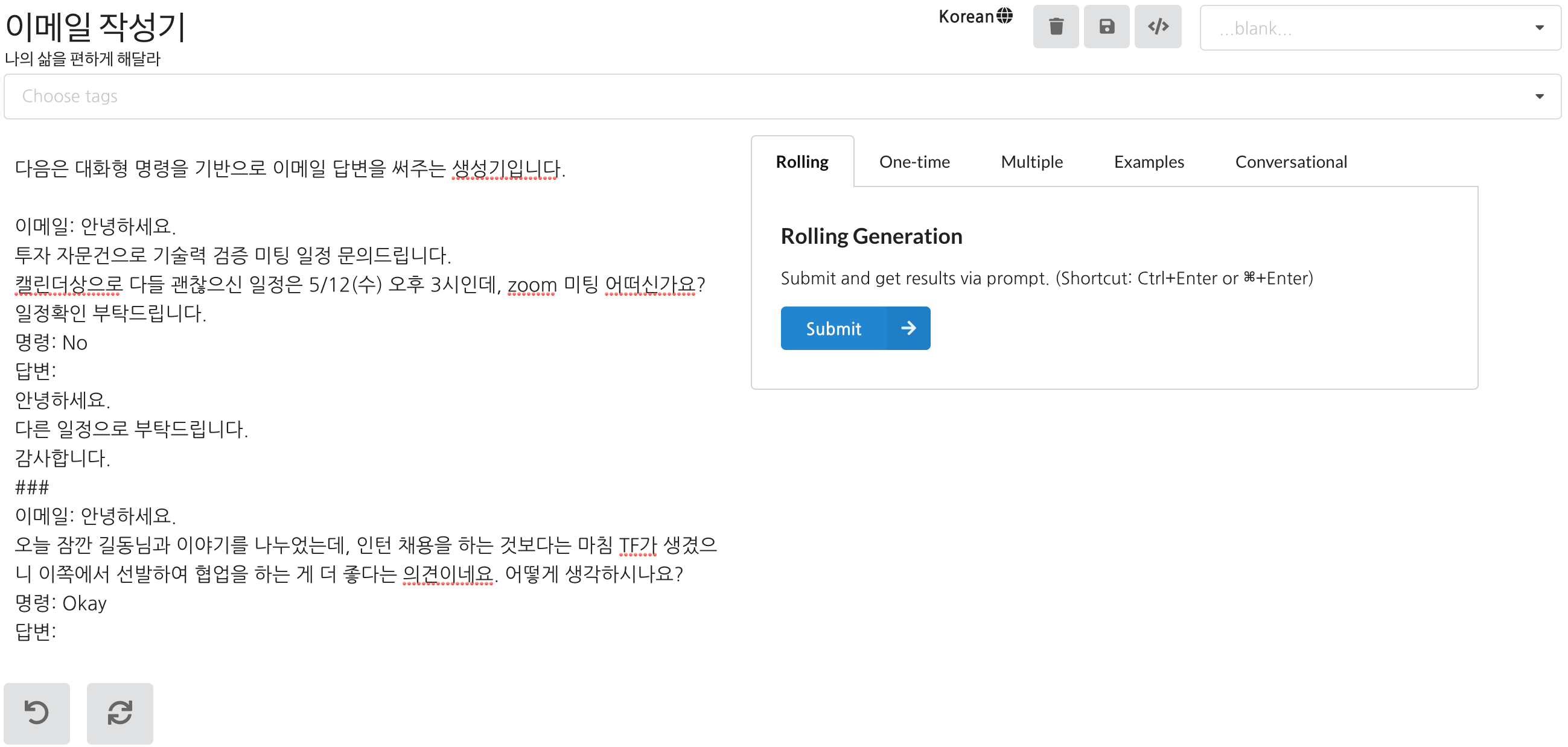}
\caption{An example interface of \modelname{} Studio.}
\label{fig:studio1}
\end{figure*}

\begin{figure*}[t] 
\centering
\includegraphics[width=1.00\textwidth]{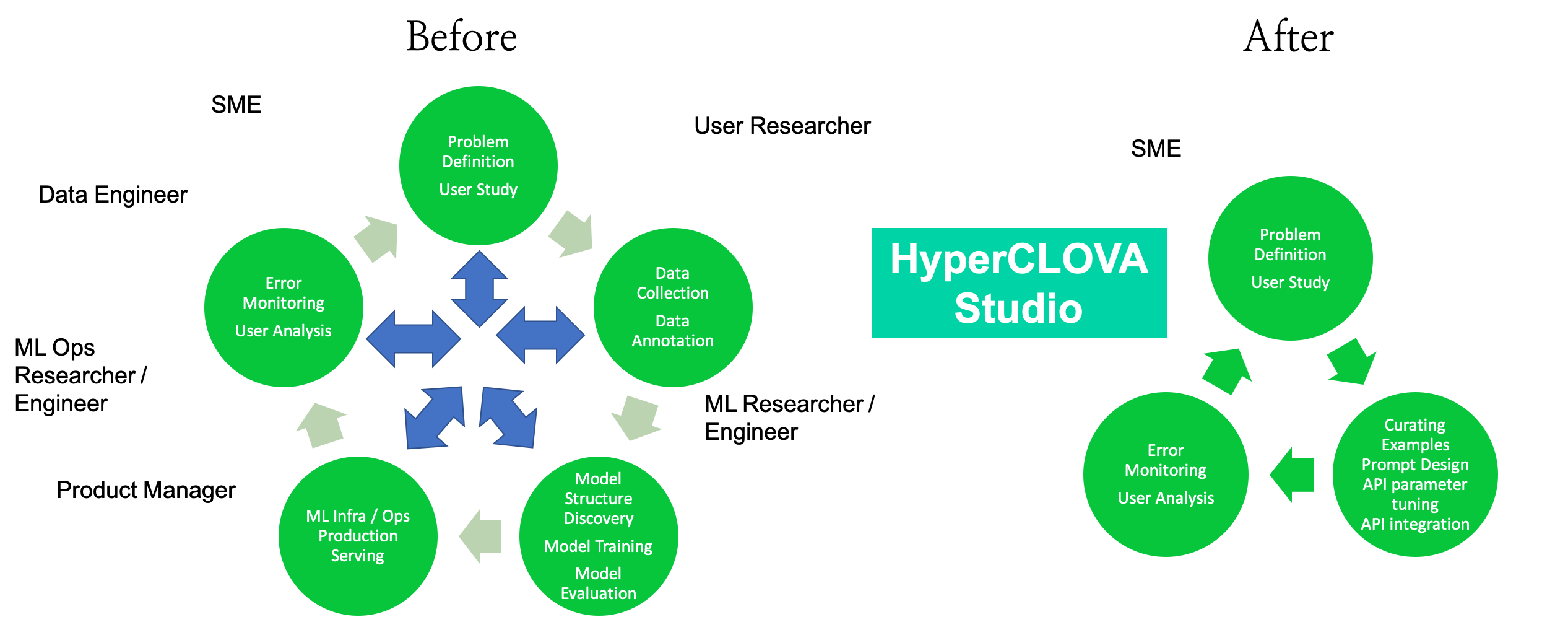}
\caption{No Code AI paradigm in \modelname{} Studio.}
\label{fig:no-code-ai-studio}
\end{figure*}

Figure \ref{fig:studio1} shows the GUI interface of \modelname{} Studio. 
Figure \ref{fig:no-code-ai-studio} illustrates No Code AI paradigm in \modelname{} Studio.

\section{Motivation of Tokenization}
\label{sec:movitation-of-tokenization}

\begin{figure*}[t] 
\centering
\includegraphics[width=1.00\textwidth]{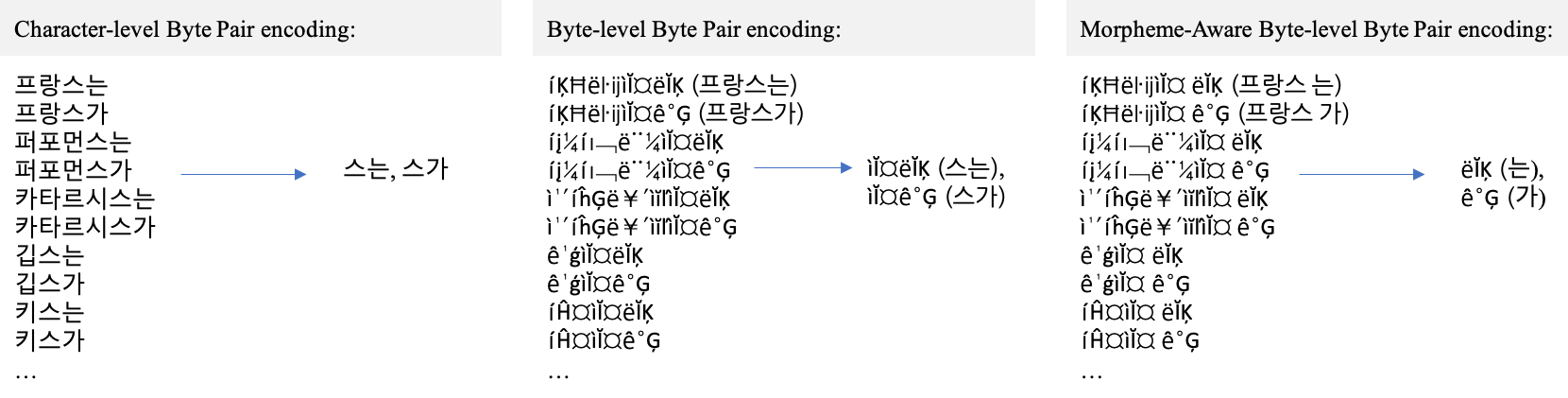}
\includegraphics[width=1.00\textwidth]{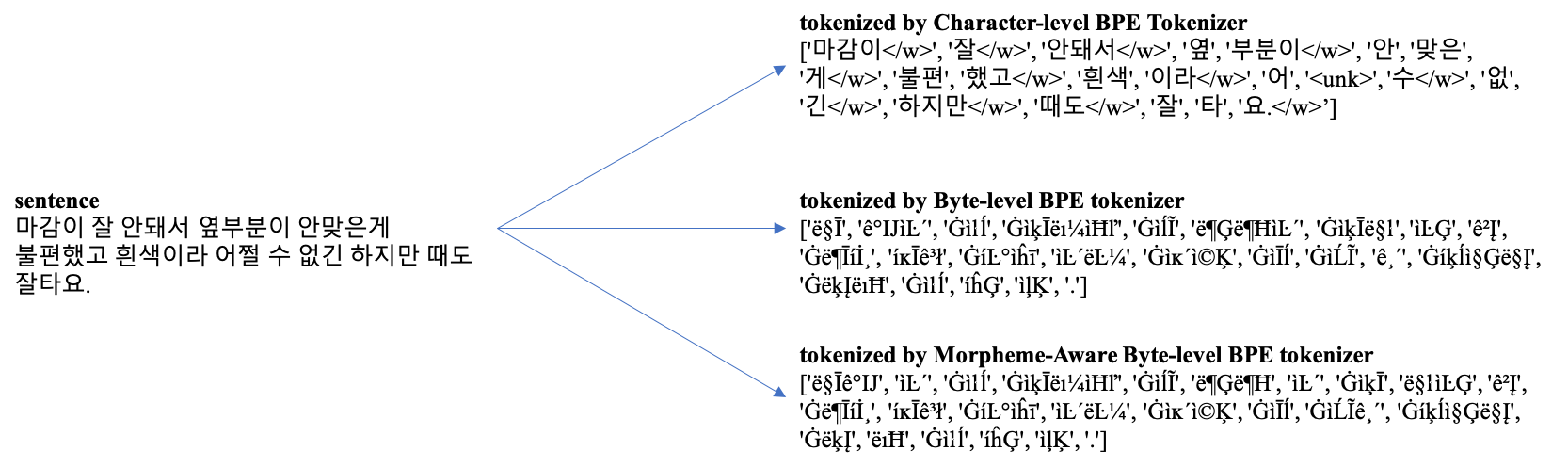}
\includegraphics[width=1.00\textwidth]{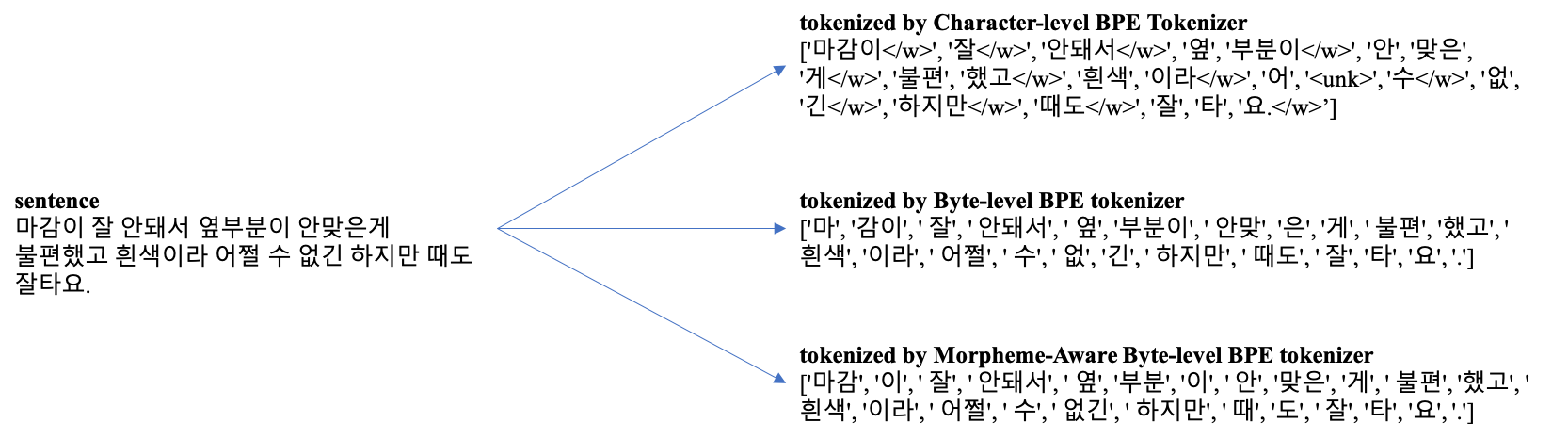}
\caption{Motivation of our morpheme-aware byte-level BPE tokenization. (Top) A conceptual example of making subword from three tokenization methods. (Middle) An example of tokenization, where subword from byte-level tokenizer is represented as a byte. (Bottom) The same example of (middle), but subword from byte-level tokenizer is represented as a character. }
\label{fig:mabbpe}
\end{figure*}

Figure \ref{fig:mabbpe} shows our motivation and importance of morpheme-aware tokenization.
Though we used an in-house morpheme analyzer, an alternative open-source morpheme analyzer like Mecab-ko\footnote{https://bitbucket.org/eunjeon/mecab-ko} can also be used.
\section{Challenges of No/Low Code AI Paradigm}
\label{sec:challenges}

Some researchers doubt the performances of GPT-3 less competitive than existing finetuning-based LMs for various downstream tasks. For example, task-specific neural structure like FiD~\cite{izacard2021leveraging} achieves state-of-the-art open-domain QA, whereas GPT-3 does not. It is still under-discovered that a prompt-based method makes large-scale LMs competitive. To resolve this problem, further discovery on general large model capability and prompt-based optimization is required.

There also exists a problem with dependency on pre-training data. If the corpus does not contain code generation, it is unfair to expect the LM generates source codes, even where a prompt-based optimization is applied. The maintainer of \modelname{} Studio may discover many requirements of users and further train corpus with common needs. To incorporate these corpora, research on pre-training under continual learning setup~\cite{bang2021rainbow} is required.

Though we mentioned No Code AI earlier, programming further the functions of \modelname{} Studio still exists for the remaining part of complete AI system. Also, knowledge of ML is still required implicitly to design an effective prompt and few-shot examples. An easier guideline for Studio and incentives on sharing user's own prompts can boost to spread the ecosystem.

In order to support a full-fledged ML development, we also need additional features for \modelname{} Studio - experimentation and user feedback. In this function, a user can easily distribute PoC service by an appropriate interface, like a text editor or messenger, and make the user can feedback on responses of \modelname{}. For example, user can rate the response of the chatbot turn by turn. 

Expensive inference or prompt-based optimization costs are still an obstacle for using large-scale LMs. However, there is a trade-off on costs between training many small-scale LMs and inferencing one large-scale LM. The outputs by one large-scale LM can also be input to small-scale LMs \cite{yoo2021gpt3mix}. Research on distilling generative transformers or energy-efficient hardware is essential for sustainability. Further discussion several issues are in the Broader Impact Statement section.

\end{document}